\documentclass[11pt,dvipsnames]{article}
\pdfoutput=1
\usepackage{coling2020}
\usepackage{times}
\usepackage{url}
\usepackage{latexsym}

\usepackage{epsf}
\usepackage{enumitem}

\usepackage{amssymb}
\usepackage{amsmath}
\usepackage{comment}
\usepackage{graphicx}
\usepackage{textcomp}
\usepackage{float}
\usepackage{siunitx}
\usepackage{subfigure}
\usepackage{setspace}
\usepackage{comment}
\usepackage{hyperref}
\usepackage{subfiles}
\usepackage[font=scriptsize]{caption}
\usepackage[justification=centering]{caption}

\usepackage{tikz}
\usetikzlibrary{calc}
\usetikzlibrary{positioning}
\tikzset{
    >=stealth,
    hair lines/.style={line width = 0.05pt, lightgray},
    true scale/.style={scale=#1, every node/.style={transform shape}},
}

\colingfinalcopy % Uncomment this line for the final submission

\usepackage[]{xcolor}
\definecolor{gray}{RGB}{219, 48, 122}

\title{Multi-Task Sequence Prediction For Tunisian Arabizi Multi-Level Annotation}

\author{Elisa Gugliotta$^{1,2,3}$, Marco Dinarelli$^{1}$, Olivier Kraif$^{2}$\\
\small{1. 
Université Grenoble Alpes - Laboratoire LIG, \emph{Getalp group}. 2. Université Grenoble Alpes - Laboratoire LIDILEM.} \\
\small{3. \emph{Sapienza} University of Rome} \\ 
         \small{elisa.gugliotta@uniroma1.it, marco.dinarelli@univ-grenoble-alpes.fr, olivier.kraif@univ-grenoble-alpes.fr}}

\date{}

\begin{document}
\maketitle
\begin{abstract}
  In this paper we propose a multi-task sequence prediction system, based on recurrent neural networks and used to annotate on multiple levels an Arabizi Tunisian corpus. The annotation performed are text classification, tokenization, PoS tagging and encoding of Tunisian Arabizi into CODA* Arabic orthography.
  The system is learned to predict all the annotation levels in cascade, starting from Arabizi input.
  We evaluate the system on the TIGER German corpus, suitably converting data to have a multi-task problem, in order to show the effectiveness of our neural architecture. We show also how we used the system in order to annotate a Tunisian Arabizi corpus, which has been afterwards manually corrected and used to further evaluate sequence models on Tunisian data.
  Our system is developed for the Fairseq framework, which allows for a fast and easy use for any other sequence prediction problem.
\end{abstract}

\section{Introduction}
\label{sec:intro}

In the last decade neural networks became the state-of-the-art models in most NLP problems.
Sequence-to-sequence models \cite{Sutskever-2014-SSL-2969033.2969173,46201}, built on top of recurrent \cite{Hochreiter-1997-LSTM,Cho-2014-GatedRecurrentUnits}, convolutional \cite{gehring2017convs2s,wu2018pay} or attentional \cite{DBLP-journals-corr-BahdanauCB14,46201} modules, and structured in encoder-decoder architectures, are currently the most effective models for NLP problems.
Neural networks have been used also for multi-task learning since early in their diffusion \cite{Collobert:ACL:2007,Collobert:2008:UAN:1390156.1390177,Collobert-2011-NLP-1953048.2078186}.

As a semitic language, Arabic has a highly inflectional and derivational  morphology, which makes Arabic processing an engaging challenge.
This morphological complexity has traditionally been handled through morphological analysers, such as BAMA \cite{buckwalter2004buckwalter}, which has been used by the Linguistic Data Consortium (LDC) to develop the Penn Arabic Treebank (PATB) \cite{maamouri2004penn}.
Recently, the number of NLP contributions to morphological analysis, disambiguation, Part-of-Speech (PoS) tagging and lemmatization has increased substantially, for both Modern Standard and Dialectal Arabic (MSA and DA, respectively).
Multitask learning was proved to be an effective way to process Arabic morphology for MSA fine-grained PoS tagging \cite{inoue2017}, as well as for DA \cite{zalmout2019joint}.
Concerning NLP applied to DA, it is possible to observe two main macro-strategies aimed at remedying the lack of data for DA:
\textbf{1.} \emph{MSA systems adaptation to DA processing}, like \cite{david2006parsing} who exploited the Penn Arabic Treebank (PATB) \cite{maamouri2004penn} and used explicit knowledge about the relation between MSA and Levantine Arabic. Instead, 
\cite{duh2005pos} built a PoS tagger for Egyptian through a minimally supervised approach by leveraging the CallHome Egyptian Colloquial Arabic corpus (ECA).
\textbf{2.} \emph{The constitution of new resources not based on MSA-DA relations}, in particular dialectal corpora, such as the Fisher Levantine Arabic Conversational Telephone Speech \cite{maamouri2007fisher}.\footnote{These resources are not freely available.}
This second strategy has been followed also collecting more \emph{ad-hoc} resources. \cite{bouamor2018madar} presented the first parallel DA corpus, collecting the dialects of 25 Arab cities, including the Tunisian dialects of Tunis and Sfax.
The MADAR corpus has been created by translating selected sentences from the Basic Traveling Expression Corpus (BTEC) \cite{takezawa2007multilingual}.
Regarding Tunisian Dialect (TD), the resource constitution strategy has been instantiated as MSA resource adaptation to the DA, e.g. building lexicons \cite{boujelbane2013building}, PoS-taggers \cite{boujelbane2014fine,hamdi2015pos}, morphological analysers \cite{zribi2013morphological} or morphological systems to disambiguate annotated transcriptions \cite{zribi2017morphological}. 
Considering the lack of freely available resources, we opted for an approach similar to the one used in \textit{Curras} Palestinian corpus collection \cite{jarrar2017curras}, which exploits MADAMIRA tools \cite{madamira}, (cf. section \ref{subsec:data}).

The development of informal online communication provided a solution to most of the data availability problems, making accessible to the scientific community a large amount of texts, both written and oral. Concerning texts written in DA, it is possible to find two main writing systems: Arabic and Latin scripts. With regard to the second one, letters are used together with digits for the encoding of those Arabic letters without correspondence in the Roman alphabet. This system is already well known as \emph{Arabizi}, or \emph{Arabish} for non-Arabic speakers.
Most of the work developed on Arabish focus on language identification \cite{Darwish14arabizidetection} and sentiment analysis \cite{duwairi2016sentiment,fourati2020tunizi}.
Several works are focused on the conversion of Arabish into Arabic script, as the Parallel Annotated Egyptian Arabish-Arabic Script SMS/Chat Corpus \cite{bies2014transliteration}. 
Transliteration has also been addressed for Tunisian Arabish \cite{masmoudi2015arabic,masmoudi2019transliteration,younes2020romanized}. 

In this paper we propose a multi-task sequence prediction system based on recurrent neural networks, that we used to annotate at multiple levels the Tunisian Arabish Corpus (TArC) \cite{gugliotta2020tarc}.
The annotation levels include tokenization, Part-of-Speech (PoS) tagging and Tunisian Arabish encoding into Arabic script.
  The system is learned to predict all the annotation levels in cascade, starting from Arabish input.
  We evaluate the system on the TIGER German corpus \cite{Brants_2004} in order to show the effectiveness of our neural architecture.
  While the purpose of this evaluation is not to improve state-of-the-art on this task, our results are comparable and sometimes better than the best published models.
  We show also how we used the system in order to annotate TArC, which has been afterwards manually corrected and used to further evaluate sequence models on Tunisian data.
  Our system is developped for \textit{Fairseq}\footnote{https://github.com/pytorch/fairseq} \cite{ott2019fairseq}, it can therefore be used for any problem involving sequence prediction.\footnote{Our system, with data used in this paper, is available at https://gricad-gitlab.univ-grenoble-alpes.fr/dinarelm/tarc-multi-task-system.
  
  The last updated version of TArC is available at https://github.com/eligugliotta/tarc.}

In the remainder of the paper we describe the TArC corpus, that we annotated with multi-level information, and we used to evaluate our neural system (in section~\ref{sec:tarc}).
In section~\ref{sec:multi-task} we describe our multi-task neural architecture for multi-level annotation, in section~\ref{sec:eval} we describe the TIGER corpus, the experimental settings, and all the results obtained with our system, on both TIGER and TArC corpora.
We conclude the paper in section~\ref{sec:conclusions}.

\section{Tunisian Arabish Multi-Level Annotated Corpus}
\label{sec:tarc}

The corpus used in this paper is the Tunisian Arabish Corpus (TArC) \cite{gugliotta2020tarc}, 
the result of a multidisciplinary work with a hybrid approach based on: 1. dialectological research questions; 2. corpus linguistics standards and 3. deep learning techniques.
TArC has been conceived with the aim to extend the dialectological investigation to the web, not only considering it as a new resource for linguistic analyses, but mainly because the object of TArC is a Computer Mediated Communication (CMC) writing system.

The gathering of CMC corpora for linguistic study purposes is a long-standing practice: as early as the 1990s, in order to study linguistic and communicational aspects, researchers began to collect corpora from mailing lists, newsgroups, electronic conferences or chat rooms \cite{yates1996oral,todla1999patterns,berjaoui2001aspects,feldweg1995sprachgebrauch}.
Nowadays, the study of CMCs is a research domain it-self, crossing various disciplines such as sociology and linguistics.
The linguistic questions related to CMC-corpora may for example concern paraverbal phenomena and the expression of emotions \cite{riordan2010emotion,tantawi2019paralinguistic}, politeness formulas and the degree of message formality \cite{brysbaert2019computer}, the effects of orality in written communication \cite{soffer2010silent}, the role of code-mixing and code-switching in mediated discourse \cite{morel2013textos,mave2018language}, 
their graphic and orthographic characteristics \cite{sullivan2017writing} (concerning Arabic).
Lastly, a lot of research deals currently with the automatic processing of such corpora \cite{lopez2018extracting,panckhurst2017linguistique}.

 Among the purposes of dialectology there is the dialect collection and description with traditional approaches: fieldwork, oral text collection and transcription, glossary building.
 We observed that in the case of Arabic varieties the descriptive landscape is made of multiple studies on single phenomena.
For this reason, we developed a resource inspired by dialectological investigation, which borrows the principles of corpus linguistics in order to guarantee representativeness, accessibility, balance and authenticity of the linguistic data \cite{Szmrecsanyi2017,wynne2005}. The data gathered in TArC, together with various metadata, takes a snapshot of Tunisian Arabish writing and its evolution over the last ten years. 
TArC is built selecting data with the following criteria: 1. \emph{text mode}: informal writing; 2. \emph{text genres}: forum, blog, social networks, rap lyrics; 3. \emph{domain}: CMC; 4. \emph{language}: Tunisian; 5. \emph{location}; 6. \emph{publication date}.
The last two items were registered via metadata extraction (publication date, user's age, gender and provenience).

The building process automation overcomes the \textit{observer's paradox} problem \cite{labov1972sociolinguistic}, an issue much discussed in dialectology \cite{boberg2018handbook}.
It also allows the reproducibility of the work, as well as the quantitative extension of an open corpus (such as TArC), which is normally difficult to ensure by dialectological research.
TArC collection has therefore been enhanced thanks to the multi-task architecture, used for a semi-automatic annotation (cf. section~\ref{sec:multi-task}) to get as close as possible to a consistent linguistic annotation \cite{wynne2005}.
The automatically generated annotations were post-edited by a linguist qualified in Arabic language and Tunisian variety, whose work was occasionally verified by native speakers.\footnote{Due to COVID-19 lockdown it was not possible to conduct the field research scheduled for March 2020.}
Such annotation work complies with both the \textit{applicative} and the \textit{analytical} purposes of a corpus. The former concerns the generation of NLP tools for the Tunisian Arabish processing. The latter is realised through the multi-functional annotation levels of TArC, which allow global and systematic studies of Tunisian variety and its Arabish encoding. 
This way, TArC usefulness returns to the dialectological area, the field in which the preliminary research questions were addressed.

TArC has been annotated with four information levels.
\textbf{1) Classification} of words in three classes:
\textit{arabizi}, \textit{foreign} and \textit{emotag}. The first class is for Tunisian and MSA words, the second one is to classify non-Arabic code-mixing; the third is used for elements as smileys or emoticons.
\textbf{2) Encoding in Arabic script} in Conventional Orthography for Dialectal Arabic (CODA*) \cite{habash2018unified}.
\textbf{3) Tokenization}, Tunisian words encoded in CODA* have been tokenized following the \emph{D3\_BWFORM} configuration scheme where basically all clitics are tokenized, including the article \cite{madamira}.
\textbf{4) Part-of-Speech} according to the \emph{PATB} guidelines \cite{maamouri2009penn}. 
All levels have been developed following the same incremental and semi-automatic procedure described in \cite{gugliotta2020tarc} for the CODAfying stage. 

\section{Multi-Task Sequence Prediction System}
\label{sec:multi-task}

There are several works about multi-task learning with neural networks for NLP problems \cite{7373350,44928}, \emph{inter alia}. Most of the time the neural architecture factorises some parameters for information that can be shared among tasks, and then uses different modules (e.g. decoders) for each task, which are learned independently.

As described in section~\ref{sec:tarc}, our goal for Tunisian Arabish data is a multi-level annotation scheme, where the different levels are potentially related. From an NLP point of view, this relations imply that some levels of annotation may help disambiguation when annotating other levels. For instance the classification information can disambiguate annotation into CODA*, tokenization and PoS tagging.
Intuitively we expected that learning tasks in chain, organised in a cascade manner in a neural network, would benefit to each other, in contrast to learning tasks individually.

\subsection{Multi-Task Neural Architecture}
\label{subsec:MTS}

We follow the intuition above and we propose a multi-task neural architecture where the different learned tasks are organised in a cascade.
The input is the Arabish text. The outputs, corresponding to the tasks to be learned, are, in this order, the classification information, the conversion into CODA* orthography, the tokenization of the \emph{CODAfied} tokens and the PoS tags.
Outputs from previous tasks are reused by the following tasks, they are thus learned jointly and interdependently.
The input is transformed into hidden context-aware representations with an encoder based on recurrent layers.
The outputs are processed by different decoders, each of them taking as input the hidden state of the encoder, and the hidden state of each of the previous decoders.
The output of each decoder is used to learn each task.

\begin{figure}
\center
\scriptsize
\begin{tikzpicture}[scale = 0.8]
	\begin{scope}[local bounding box=net]
	
	\node (o) at (0,7.5) {$\hat{o}_3$};
	
	\node[draw, rectangle, rounded corners=1pt, scale=1.2] (d3) at (0,6.0) {\emph{$\text{Decoder}_3$}};
	
	\node[draw, rectangle, rounded corners=1pt, scale=1.2] (d2) at (0,4.0) {\emph{$\text{Decoder}_2$}};
	
	\node[draw, rectangle, rounded corners=1pt, scale=1.2] (d1) at (0,2.0) {\emph{$\text{Decoder}_1$}};
	
	\node[draw, rectangle, rounded corners=1pt, scale=1.2] (e) at (0,0.0) {\emph{Encoder}}; % traits contextuels
	
    \node (i) at (0,-1.0) {x};
    
    % Central connections between 2 nodes
    \draw[->, thick] (i) -- (e);
    \draw[->, thick, dashed] (e) -- (d1);
    \draw[->, thick, dashed] (d1) -- (d2);
    \draw[->, thick, dashed] (d2) -- (d3);
    \draw[->, thick] (d3) -- (o);

    % Skip connections
    \draw[->, thick, dashed] (d1.north east) to[bend right=70] (d3.south);
    
    \draw[->, thick, dashed] (e.north west) to[bend left=70] (d3.south);
    \draw[->, thick, dashed] (e.north west) to[bend left=70] (d2.south);

    % Decoder outputs
    \node (o1) at (-3.0,4.0) {$\hat{o}_1$};
    \node (o2) at (-3.0,6.0) {$\hat{o}_2$};
    
    % Losses
    \node (L1) at (-4.0,5.0) {$\mathcal{L}_1(o_1, \hat{o}_1)$};
    \node (L2) at (-3.0,7.0) {$\mathcal{L}_2(o_2, \hat{o}_2)$};
    \node (L3) at (-2.0,8.0) {$\mathcal{L}_3(o_3, \hat{o}_3)$};
    \node[draw, circle, scale=0.8] (L) at (-4.0,8.0) {$+$};
    \node (LL) at (-4.0,9.0) {$\mathcal{L}$};
    
    % Connection to outputs
    \draw[->, thick] (d1.north) -- (o1);
    \draw[->, thick] (d2.north) -- (o2);
    
    % Connection to Losses...
    \draw[->, thick] (o1.north west) -- (L1.south);
    \draw[->, thick] (o2.north) -- (L2.south);
    \draw[->, thick] (o.west) -- (L3.south east);
    % ...Losses to sum
    \draw[->, thick] (L1.north) -- (L.south);
    \draw[->, thick] (L2.north) -- (L.south east);
    \draw[->, thick] (L3.west) -- (L.east);
    % ...sum to final Loss
    \draw[->, thick] (L.north) -- (LL.south);

    % Hidden states
    \node (hE1) at (0.3,1.0) {$h_E$};
    \node (hE2) at (-1.2,1.3) {$h_E$};
    \node (hE3) at (-2.2,2.0) {$h_E$};
    \node (h11) at (0.3,3.0) {$h_1$};
    \node (h12) at (1.8,3.4) {$h_1$};
    \node (h2) at (0.3,5.0) {$h_2$};

    \end{scope}

\end{tikzpicture}
    \caption{A high-level schema of our multi-task neural system}\label{fig:multitasksystem}
\end{figure}
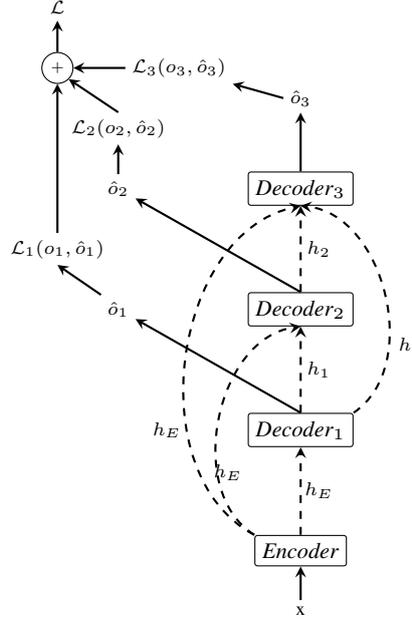

More formally, let the task $i$ be represented by the model $M_i(x, H_{i})$, with $x$ the input (Arabish text representations), and $H_{i}$ the list of hidden states from the previous models, plus the current model's hidden state $h_i$. Each model $i$ generates an output $\hat{o}_i$ and a hidden state $h_i$. $\hat{o}_i$ is the predicted output, which is used to learn the task $i$ by computing a loss $\mathcal{L}_i(o_i, \hat{o}_i)$ comparing $\hat{o}_i$ to the expected output $o_i$.
Internally the global model $M$ is made of an \emph{Encoder} and $I$ decoders  \emph{Decoder}$_i$, with $i = 1 \dots I$. The list $H_{i}$ includes both the encoder hidden state $h_E$ and the decoders hidden states $h_1...h_i$.
An high-level schema of this architecture, with the flow of information for three tasks ($I = 3$), is presented in figure~\ref{fig:multitasksystem}.
All the tasks are learned jointly by minimising the global loss $\mathcal{L} = \sum_i \mathcal{L}_i(o_i, \hat{o}_i)$, on top of the circled $+$ in the schema (Figure \ref{fig:multitasksystem}).

Like in the original sequence-to-sequence model based on an attention mechanism \cite{DBLP-journals-corr-BahdanauCB14}, each decoder attends to encoder and decoder's hidden state information with an attention mechanism.
The decoder \emph{Decoder}$_i$ has therefore $i$ different attention mechanisms, one for attending encoder's information, and one for each previous decoder's hidden state. The queries for the attention mechanisms are always the \emph{Decoder}$_i$'s hidden states, while keys and values are the encoder and previous decoder hidden states. The attention vectors computed by the attention mechanisms are \emph{simply} summed together to generate the final state, used to predict the next output.\footnote{We note that we have been testing also gating mechanisms to \emph{blend} the outputs of the attention mechanisms like in \cite{miculicich-etal-2018-document}, but this always gave worse results than the sum.}
 
\section{Evaluation}
\label{sec:eval}

\subsection{Data}
\label{subsec:data}

In order to evaluate our multi-task system, we used two different corpora. One is TArC, described in section~\ref{sec:tarc}, the other is the German TIGER corpus \cite{Brants_2004}.

\textbf{TArC corpus} has been initially collected from forums, social media and blogs, for a total of 32~062 words, and recently extended to 43~313 words by adding the text type of rap lyrics.
In order to better organise the automatic annotation and the manual-correction stages, we split the initial corpus into blocks of roughly 6~500 tokens. Statistics of TArC are presented in table~\ref{tab:ArabiziStats}.
The initial model, used to bootstrap the corpus annotation, has been trained using 2000 sentences from the Tunisian MADAR corpus.
MADAR data are well-formed texts encoded in Arabic script, this avoid any code-switching and spelling inconsistency.
We processed MADAR data using the MADAMIRA tool \cite{madamira}.\footnote{Version used: MADAMIRA 2.0. D3\_BW* schemes \cite{habash2010introduction}.}, producing tokenization and PoS tags.
After a manual correction, we obtained the first TArC training block for starting the annotation procedure.

\textbf{The German corpus TIGER}
\cite{Brants_2004} is annotated with rich morpho-syntactic information. These include PoS tags, but also gender, number, cases, and other inflection information, as well as conjugation information for verbs.
The combination of all these components constitutes the output labels.
We used the same data split used in \cite{Lavergne17learning}. Statistics of this corpus are given in table~\ref{tab:TIGERStats}.

\begin{table}[t]
    \begin{minipage}{0.5\linewidth}
        \centering
        \scriptsize
        \begin{tabular}{|l|rr|rr|rr|}
        \hline
          & \multicolumn{2}{c}{Training} & \multicolumn{2}{|c|}{Dev} & \multicolumn{2}{c|}{Test}\\
          \hline
          \# sentences     &\multicolumn{2}{c}{40~472} &\multicolumn{2}{|c|}{5~000}&\multicolumn{2}{c|}{5~000} \\
        \hline
          \hline
          & \multicolumn{1}{c}{Words} & \multicolumn{1}{c|}{Labels} &  \multicolumn{1}{c}{Words} & \multicolumn{1}{c}{Labels} &
        \multicolumn{1}{|c}{Words} & \multicolumn{1}{c|}{Labels} \\
        \hline
        \# tokens          & 719~530 & -- & 76~704 & -- & 92~004 & -- \\
        dictionary         &  77~220 &     681 &    15~852 &    501 & 20~149 &     537 \\
        OOV\%   & --     & --     &  30,90  & 0,01  &  37,18  &  0,015  \\
        \hline
        \end{tabular}
        \caption{Statistics of the German corpus TIGER}
    \label{tab:TIGERStats}
    \end{minipage}
    \begin{minipage}{0.5\linewidth}
        \centering
        \scriptsize
        \begin{tabular}{|l|rr|rr|rr|}
        \hline
          & \multicolumn{2}{c}{Sentences} & \multicolumn{3}{|c}{Words}
          & \multicolumn{1}{c|}{}\\
        \hline
          \ Total
          &\multicolumn{2}{c}{4121} &\multicolumn{3}{|c}{32~062}&\multicolumn{1}{c|}{} \\
        \hline
          \hline
          &  &   & \multicolumn{1}{c|}{\emph{arabizi}} & \multicolumn{1}{c|}{\emph{foreign}} &
        \multicolumn{1}{c}{\emph{emotag}} & \multicolumn{1}{c|}{} \\
        \hline
        forum &  & 756  & \multicolumn{1}{c|}{6039} & \multicolumn{1}{c|}{5856} &
        \multicolumn{1}{c}{14} & \multicolumn{1}{c|}{} \\
        social &  & 3146  & \multicolumn{1}{c|}{11~843} & \multicolumn{1}{c|}{3614} &
        \multicolumn{1}{c}{587} & \multicolumn{1}{c|}{} \\
        blog &  & 219  & \multicolumn{1}{c|}{3763} & \multicolumn{1}{c|}{343} &
        \multicolumn{1}{c}{3} & \multicolumn{1}{c|}{} \\
        \hline
        \end{tabular}
        \caption{Statistics of the already annotated part of TArC}
    \label{tab:ArabiziStats}
    \end{minipage}
\end{table}

\subsection{Settings}
\label{subsec:settings}

\subsubsection{Data Pre-processing}

We first describe some data pre-processing performed on both corpora, in order to better exploit the small amount of data in TArC, on one side; on the other side, we performed a similar pre-processing on the TIGER corpus, in order to have similar experimental settings and therefore be able to validate the multi-task model with results comparable with the literature.

The TIGER corpus has been used as a benchmark for our multi-task system, before applying it to TArC.
Since TIGER data are not natively multi-task, we re-organised TIGER labels in two parts: the first consisting of the PoS core-tag only, the second consisting of the whole label.
For example, given the label \emph{ADJA.PoS.Nom.Sg.Masc}\footnote{The different pieces stand for adjective, possessive, nominative, singular and male, respectively.}, we take the PoS tag \emph{ADJA} as a first level of information, and the whole label as a second level. This simple pre-processing allows to have two tasks to learn with our system: a coarse and a fine-grained morpho-syntactic tagging,
where the second task, more complex, can be learned using also the information of the first, which is simpler.

In order to reduce data sparsity in TArC, we performed sequence prediction at each annotation level using sub-token units, except for the classification level.
Sub-token units are characters for Arabish, \emph{CODAfied} tokens and tokenization levels.
For the PoS tags we performed an \textit{ad-hoc} split into coarser units.
The PoS tags annotated in TArC follow the LDC guidelines described in \cite{maamouri2009penn}.\footnote{In the concatenation style we used "-" and the square brackets, to better manage the information through our model.} 
The tags contain rich information, like for the TIGER labels, describing the morphological structure of tokens.
For instance the tag \emph{PV-PVSUFF\_SUBJ:3MS+[PREP+PRON\_2S]PVSUFF\_IO:2S}, contains information about a verb with inflectional morphology (\emph{PV-PVSUFF\_SUBJ:3MS}), plus information on a pre-pronominal enclitic group attached to the verb (\emph{PREP+PRON\_2S}).
This group contains also an indirect object in suffix form (\emph{PVSUFF\_IO:2S}). 
Each of these 3 macro components contains person features, \emph{3MS} for the verb, \emph{2S} for the enclitic pronoun, and \emph{2S} for the indirect object suffix.

Quite intuitively, such complex tags, taken as a whole, are very rare in the data. Indeed more than half of them occur only once in our data.\footnote{More precisely, 423 PoS tags out of 776 in the dictionary, that is 54.9\%, occur only once.}
However their components are quite common (e.g. \emph{PV, PVSUFF, SUBJ, 3, M, S} and so on).
For this reason we split the tag above into a sequence of components like: \emph{PV, PVSUFF, \_SUBJ, :3, @M, @S, +, [, PREP, +, PRON,\_2, @S, ], PVSUFF, \_IO, :2, @S}. Symbols like \emph{@} are added for the post-processing phase to correctly reconstruct the whole tag. For the same reason, each time a tag is split in this way, the components are wrapped with start and end markers \emph{<SOT>, <EOT>} (for Start and End Of Token).
A whole tag sequence, associated to an input sentence, is made by concatenating the sequences resulting from the split of each tag.
The same start and end markers are used also for the other annotation levels, which are split into single characters, so that the model can learn itself that each token in the input sequence corresponds to one token in all the annotation levels.

In order to have the same settings for TArC and TIGER data, we split the input tokens in the TIGER data into characters, adding the start and end markers. The labels are left unchanged, beyond artificially creating 2 label levels to test our multi-task system (we actually performed experiments also splitting TIGER labels into components, cf. table~\ref{tab:TIGERResults} and \ref{tab:TIGERTESTResults}).

The TArC classification level was added first. This was done using a character-level model pre-trained exploiting: i) the Hussem Ben Belgacem’s French dictionary, consisting of 336,531 tokens.\footnote{https://github.com/hbenbel/French-Dictionary (last access on 15/09/2020).}, and ii) a Tunisian Arabish dictionary of 100,936 tokens, resulting from the merge of the TUNIZI Sentiment Analysis Tunisian Arabic Dataset \cite{fourati2020tunizi}\footnote{https://github.com/chaymafourati/TUNIZI-Sentiment-Analysis-Tunisian-Arabizi-Dataset (last access on 15/09/2020).} and the TLD dataset \cite{younes2015constructing}.

In order to obtain an \textit{emotag} dictonary, we extracted all the smileys and emoticons from the Arabish dictionary above.
Once the model was pre-trained on the above data, it was possible to apply also to this annotation level the semi-automatic and incremental annotation procedure used in \cite{gugliotta2020tarc}.
At the end of the procedure, the model reached 97\% of accuracy. All data were manually checked and corrected.

\subsubsection{Model Settings}

Concerning model settings, we note that encoder and decoders in our multi-task neural models are all LSTM \cite{Hochreiter-1997-LSTM}.\footnote{The system is however generic, and potentially any kind of encoder and decoder available in Fairseq may be used. We are currently working on adding the use of Transformer encoder and decoders \cite{46201}.}

An optimisation of hyper-parameters like learning rate, dropout ratio \cite{JMLR:v15:srivastava14a}, layer size, etc. has been performed on development data of TArC. For experiments on TIGER the same hyper-parameters have been used. The goal here is not to obtain the best absolute results on this task, it is to show that our system is competitive enough to be used safely on unpublished data.
Such hyper-parameter optimal values resulted in: $5E^{-4}$ for learning rate, 0.5 for dropout ratio (at all layers, including embeddings), 5.0 for gradient clipping \cite{DBLP:journals/corr/abs-1211-5063}, 256 for both embeddings and hidden layer size (for all layers). We share all embeddings, at input and output layers, and in encoder and decoders.

The loss functions used in all our experiments, for all the decoder outputs (see $\mathcal{L}$, $\mathcal{L}_1$, etc. in section~\ref{subsec:MTS}), are the cross-entropy loss.
All models are learned with an ADAM optimiser \cite{kingma2014method} with default parameters.
Model's outputs are evaluated with the accuracy, after applying post-processing to reconstruct original tokens.
This means that if a single character or component in a token is wrong, the token is considered wrong in the accuracy.

\subsection{Results}
\label{subsec:results}

We present first results obtained on the corpus TIGER.
We remind that we artificially performed multi-tasking on TIGER by isolating the core-tag from its features for each morpho-syntactic tag, and using the core-tag and the whole one as separated output to be predicted (see section~\ref{subsec:settings}).

The first set of experiments was performed to choose the optimal number of layers in each decoder of our multi-task system.
Results are shown in table~\ref{tab:TIGERResults}, the two tasks are \textbf{PoS}, for core-tags only, and \textbf{MORPHO} for core+feature tags.
The results of both tasks show that the model performs at best with 3 layers in each decoder, though the gain with respect to the other choices is small.
Despite the gain is small, we observed consistently the best results, both in terms of accuracy and loss values, and on both corpora, with 3 layers.

In the table~\ref{tab:TIGERResults} we show also the comparison of our results with the literature. To the best of our knwoledge the best results on the corpus TIGER have been published in \cite{DinarelliGrobol-Seq2BiseqTransformer-2019}, which improved previous state-of-the-art of \cite{Lavergne17learning}.
Our results are comparable with the state-of-the-art, even slightly better on morpho-syntactic tagging, \textit{Dev data}.
We would like to insist on the fact that experiments on TIGER have been performed not with the goal to improve the state-of-the-art, but only for validating our multi-task system for performing multi-level annotation of TArC as multi-tasking.
In this respect, the model used in \cite{DinarelliGrobol-Seq2BiseqTransformer-2019} is quite sophisticated, it performs sequence labelling exploiting both token and character information on the input side, and performing bidirectional decoding on the output side. Our model performs decoding at character-level only, though using several layers over 2 tasks.
Beyond this comparison, we consider our results on the TIGER corpus satisfactory for a multi-task setting.

The last 3 lines of table~\ref{tab:TIGERResults} and \ref{tab:TIGERTESTResults} show results on TIGER Dev and Test data, respectively. In these experiments we compare models learned for decoding label components, instead of whole labels, using character-level input (\textit{Char decoding}), models learned with whole tokens on input and output side (\textit{Token decoding}), and models combining both information, but learned from whole-token tag sequences (\textit{Token+char decoding}). As we can see, \textit{Char decoding} setting is by far the most effective. Combining token and character level information largely improves the \textit{Token decoding} setting, but it is still much less effective than the \textit{Char decoding} setting.

It could be interesting to observe which gain can be achieved with a multi-task model, e.g. on PoS tagging,
with respect to a mono-task sequence-to-sequence model on the same task. In order to show such gain, we performed an experiment of PoS tagging with our multi-task system in a mono-task setting, with the same experimental settings.
We compare this result with the multi-task counter-part in table~\ref{tab:TIGERResults}. The two results are shown in table~\ref{tab:POSComparison}.
As we can see, a substantial gain can be achieved performing PoS tagging as part of a multi-task setting. Even if, when learned for multi-tasking, PoS tagging is the first task and so it cannot exploit information coming from preceding tasks, the gain is given by the back-propagation of the morpho-syntactic tagging error through the whole network.
Once again, results are obtained decoding at character level only for keeping the same experimental settings as for the TArC.

\begin{table}[t]
    %\caption{}
%\begin{center}
    \begin{minipage}{0.5\linewidth}
        \centering
        \scriptsize
        \begin{tabular}{|l|cc|}
            \hline
            \multicolumn{3}{|c|}{Corpus: TIGER Dev data} \\
            \hline
            \multicolumn{3}{|c|}{\textbf{Best results}} \\
            & \multicolumn{1}{c}{\textbf{PoS}} & \multicolumn{1}{c|}{\textbf{MORPHO}}\\
            \hline
            \cite{DinarelliGrobol-Seq2BiseqTransformer-2019}   & \multicolumn{1}{c}{98.37\%}	& \multicolumn{1}{c|}{93.94\%} \\
            \hline
            \hline
            \multicolumn{3}{|c|}{\textbf{Our results}} \\
            \textbf{Model} & \multicolumn{2}{c|}{\textbf{LSTM}} \\
            \hfill \textbf{Task} & \textbf{PoS} & \textbf{Morpho} \\
            \hline
            \hline
            1 Enc + 1 Dec layers	    & 97.83\%	& 93.16\% \\
            2 Enc + 2 Dec layers	    & 98.16\%	& 93.58\% \\
            3 Enc + 3 Dec layers		& 98.30\%	& 94.10\%  \\
            \hline
            \hline
            Char decoding		        & 98.30\%	& 94.10\%  \\
            Token decoding		        & 96.21\%	& 86.89\%  \\
            Token+char decoding		        & 98.11\%	& 90.70\%  \\
            \hline
        \end{tabular}
        \caption{Summary of results, in terms of accuracy, obtained on  
        the TIGER development data set with the Tarc Multi-Task system.}
        \label{tab:TIGERResults}
    \end{minipage}
    \begin{minipage}{0.5\linewidth}
        \centering
        \scriptsize
        \begin{tabular}{|l|cc|}
            \hline
            \multicolumn{3}{|c|}{Corpus: TIGER Test data} \\
            \hline
            \multicolumn{3}{|c|}{\textbf{Best results}} \\
            & \multicolumn{1}{c}{\textbf{PoS}} & \multicolumn{1}{c|}{\textbf{MORPHO}}\\
            \hline
            \cite{DinarelliGrobol-Seq2BiseqTransformer-2019}   & \multicolumn{1}{c}{97.74\%}	& \multicolumn{1}{c|}{91.86\%} \\
            \hline
            \hline
            \multicolumn{3}{|c|}{\textbf{Our results}} \\
            \textbf{Model} & \multicolumn{2}{c|}{\textbf{LSTM}} \\
            \hfill \textbf{Task} & \textbf{PoS} & \textbf{Morpho} \\
            \hline
            Char decoding		        & 97.44\%	& 91.81\%  \\
            Token decoding		        & 94.44\%	& 83.37\%  \\
            Token+char decoding		        & 97.25\%	& 87.87\%  \\
            \hline
        \end{tabular}
        \caption{Summary of results, in terms of accuracy, obtained on  
        the TIGER test data set with the Tarc Multi-Task system.}
        \label{tab:TIGERTESTResults}
        
        \vspace{0.5em}
        \scriptsize
        \begin{tabular}{|l|c|}
            \hline
            \multicolumn{2}{|c|}{Corpus: TIGER Dev data} \\
            \hline
            \multicolumn{2}{|c|}{\textbf{PoS tagging results}} \\
            \textbf{Model} & \multicolumn{1}{c|}{\textbf{LSTM}} \\
            \hline
            \hline
            Mono-task       &   95.66\% \\
            Multi-task		&   98.30\% \\
            \hline
        \end{tabular}

        \caption{Comparison of results of PoS tags decoding from source characters, on the TIGER development data with mono-task and multi-task models.}
        \label{tab:POSComparison}
    \end{minipage}
\end{table}

Experiments on TArC are divided in two phases, corresponding to two annotation phases: the first concerns the Arabish conversion into Arabic script. The second phase consists in classification of each token in \textit{arabizi}, \textit{foreign} or \textit{emotag} classes, together with tokenization of Arabic-encoded tokens, and PoS tagging.
Each phase was performed with a semi-automatic procedure, where a model was trained on a first block of data.
Such model was used to annotate another block of data.
This was then manually corrected and added to the training data. A new model was trained and used to annotate a new block. This procedure was iterated up to the annotation of the full corpus (32~062 tokens).

For the first phase of the annotation (Arabic script encoding only) we used the mono-task sequence-to-sequence model of \cite{DinarelliGrobol-Seq2BiseqTransformer-2019}.
Indeed the Arabic script encoding of tokens is the most costly and difficult phase, so we thought it could be easier to have it first, annotating the other levels afterwards.
The Arabic script encoding accuracy of the model was below 70\% for the first block. This still allowed the annotator to correct the block 3 times faster than if the block was annotated from scratch.
For the following data blocks, accuracy of the model increased progressively, up to roughly 76\% for the fourth block.
At this point we started the second phase, which included the annotation of the fifth and last block with encoding conversion.

In the second phase, we repeated the iterative semi-automatic annotation procedure of the first phase for the classification, tokenization and PoS tagging levels. These were performed with the multi-task system.
The first model for bootstrapping the annotation procedure was trained on a part of the MADAR data \cite{bouamor2018madar} consisting of roughly 12,000 tokens (~2,000 sentences).
These data were annotated with tokenization and PoS information using MADAMIRA as explained in section~\ref{subsec:data}, and then manually corrected. The classification information was added manually, which was trivial since all tokens belong to the \textit{arabizi} class in this data.
The model trained on MADAR data has been used to annotate the first block of TArC, which is the step 0 of the iterative procedure. In the following 3 iterations, the MADAR data were used together with the TArC blocks already manually corrected.
The input for these 3 steps was thus the \emph{CODAfied} Tunisian.
Exploiting MADAR was only possible up to the 4th block, since the blocks after the fourth were not already provided with \emph{CODAfied} tokens (see the 1st annotation phase above).
However, we planned to add all the annotation levels to the 5th block, including the encoding in Arabic script level, with the multi-task system.
The 5th block was thus annotated using only TArC four blocks in Arabish as training data.
At each iteration step, the Arabish data were split randomly into train and validation (dev) sets, so that the dev set is representative of the whole data at each iteration.\footnote{In this respect, we note that data in different blocks are heterogeneous, as they are not all from the same source. Hence keeping the same dev data set for all the iterations would not be representative.}

We report the results on the 3 tasks of the first 4 steps, where the input was \emph{CODAfied} Tunisian, and the results on the 4 tasks of the following steps, where the input was Arabish, in table~\ref{tab:MADATarc_AccIterativeResults}.
The tasks are indicated in the table with \textbf{Class} for classification, \textbf{Arabic} for Arabic script encoding, \textbf{Token} for tokenization, and \textbf{PoS} for PoS tagging, respectively. In the column \textbf{``Train. tokens''} of the table we report the number of training tokens for each step. Between parenthesis, when this is meaningful, we also report the number of training tokens coming from TArC (the remainder is from the MADAR corpus).

In table~\ref{tab:MADATarc_AccIterativeResults}, \emph{Step0} is the bootstrapping step, where the model is trained on MADAR data only. Results are on a randomly chosen dev data set consisting of 15\% of the whole data set.
Starting from \emph{Step1}, the dev data set is a 15\% random split of the TArC data only, as we are interested in the effectiveness of our multi-task system on spontaneous and informal writing data for annotation purposes.

Results in table~\ref{tab:MADATarc_AccIterativeResults} prove that the multi-task system is effective also on TArC, especially taking into account the small amount of data available for training the models. The classification task (\textbf{Class}) is quite well solved, as at best the model, when evaluated on TArC text, is over 97\% of accuracy. Results for tokenization (\textbf{Token}) are also satisfactory, in particular at step 3, where the model is over 91\% of accuracy.
Results on PoS tagging (\textbf{PoS}) are quite lower with respect to the other tasks, but we note that this task is the most difficult, among the 3 of the first 4 steps.
Indeed, classification only consists in associating to each token one of the 3 classes \textit{arabizi}, \textit{foreign} or \textit{emotag}. 
The tokenization task consists in splitting a \emph{CODAfied} token into its components with some orthographical transformations, input and output script is thus the same, the model needs to learn the splitting.
In contrast, PoS tagging is a conversion from Arabic characters into PoS components.

As we have explained in section~\ref{subsec:settings}, PoS tags are quite complex, and splitting them into components allows to mitigate the problem of data sparsity. Moreover, accuracy is computed after post-processing, that is after PoS tags have been reconstructed from components. A single mistake on a component results in a wrong tag, affecting the accuracy.
Taking all of that into account, we consider the best PoS tagging result of 76.38\% of accuracy as an acceptable result.

In table~\ref{tab:MADATarc_AccIterativeResults} we observe a substantial drop of results from step 0 (where the model is evaluated on the MADAR dev set) to step 1 (where the model is evaluated on the Arabish dev set only).\footnote{All MADAR tokens are classified as \emph{arabizi}, it is thus normal that the model gets almost perfect result in classifying it.} This is not surprising, as MADAR is made of morphosyntactically well-formed text, while TArC is made of CMC spontaneous texts.
This behaviour is useful to explain the difference of results between step 3 and step 4 and 5.
Beyond that, the increased amount of TArC data with respect to MADAR data through steps 1 to 3, allows to improve results obtained on the MADAR data (\emph{Step0}).

\begin{table}[t!]
\begin{center}
\scriptsize
    \begin{tabular}{|l|c|cccc|}

        \hline
        & \textbf{Train. tokens} & \multicolumn{4}{c|}{\textbf{LSTM}} \\
        \textbf{Task} & & \textbf{Class} & \textbf{Arabic} & \textbf{Token} & \textbf{PoS} \\
        \hline
        \hline
        \multicolumn{6}{|c|}{Corpus: MADAR} \\
        \hline
        Step0 & 12~391  & 99.83\%   &   -   &   88.83\%   &   72.71\% \\
        \hline
        \hline
        \multicolumn{6}{|c|}{Corpus: MADAR+TArC} \\
        \hline
        Step1 & 17~261 (4~870)  & 92.69\%   &   -   &   77.66\%   &   59.56\% \\
        Step2 & 22~173 (9~780)  & 97.21\%   &   -   &   87.53\%   &   74.30\% \\
        Step3 & 27~270 (14~870)  & 96.69\%   &    -   &   91.47\%   &   76.38\% \\
        \hline
        \hline
        \multicolumn{6}{|c|}{Corpus: TArC} \\
        \hline
        Step4 & 22~150  &   96.83\% &   75.30\% &   73.38\% &   69.76\% \\
         Step5 & 27~435  &   97.17\% &   75.08\% &   73.07\% &   66.24\% \\
         \hline
         Step4$_{\text{smart-init}}$ & 22~150  &   95.91\% &   76.55\% &   74.96\% &   72.57\% \\
         Step5$_{\text{smart-init}}$ & 27~435  &   97.08\% &   77.83\% &   75.69\% &   69.76\% \\
        \hline
    \end{tabular}
\end{center}
\caption{Summary of results, in terms of accuracy, obtained on the TArC data at the different steps of the iterative procedure for semi-automatic annotation of the corpus. The tasks are indicated with \textbf{Class} for classification, \textbf{Arabic} for Arabic script encoding, \textbf{Token} for tokenization, and \textbf{PoS} for PoS tagging.}
\label{tab:MADATarc_AccIterativeResults}
\end{table}

Results in table~\ref{tab:MADATarc_AccIterativeResults} drop again between steps 3 and 4. We remind that at step 3, data blocks from 1 to 3, plus the MADAR data, are used for training the model, a 15\% split of the TArC data are used for validation, and the model is used to annotate the fourth data block. At step 4 only TArC data are used for training, again a 15\% split is used for validation, and the fifth block is annotated. At this step an additional task is performed: encoding of Arabish into CODA*.

As we can see in table~\ref{tab:MADATarc_AccIterativeResults}, all results except for classification, substantially dropped.
This is due to having an additional task with respect to the previous steps, and thus an additional decoder in the system, and to the use of a smaller training set.
We note however this drop is similar to the one between steps 0 and 1. We conclude thus that MADAR well-formed texts have a  positive effect on learning spontaneous Arabish text.
It is interesting to observe that the drop in PoS tagging results with respect to tokenization, at steps 4 and 5, is much smaller than the drop at steps 1 and 3. This suggests to improve Arabish \emph{CODAfication} results, which may be achieved by adding Arabish encoding to MADAR.
Results on the step 5 are similar to step 4. This is not surprising as well, since data in the block 5 have a different style, coming from a different source (blogs). This balances the increased amount of data for training the model.

In order to exploit the MADAR data also at steps 4 and 5, we designed an \emph{ad-hoc} parameter initialisation using the model trained at step 0. Note that such model has a different architecture as MADAR is in Arabic script, it doesn't contain Arabish.\footnote{This resulted in a quite task-specific parameter initialisation.}
Results obtained with this initialisation are reported in the last lines of table~\ref{tab:MADATarc_AccIterativeResults} marked as \emph{smart-init}. As we can see, except for the classification task which is biased by the fact that in MADAR all tokens are in the \emph{arabizi} class, all other task results improved with respect to step 4 and 5 without pre-initialisation.

\section{Conclusions}
\label{sec:conclusions}

We presented a multi-task sequence labeling system based on recurrent neural networks, developed for the Fairseq framework and used to annotate TArC on multiple levels. The annotation levels provided are: classification, tokenization, PoS tagging and encoding of Tunisian Arabish into Arabic script, according to  CODA*. We described the annotation procedure, after showing the effectiveness of our neural architecture with an evaluation on the TIGER German corpus. As a next stage we plan to expand TArC quantitatively to improve the results and its usability in linguistics and NLP fields. Future work includes qualitative extension through the addition of further annotation levels, such as lemmatization.
%, following the same semi-automatic procedure described in this paper.
% We are also interested in applying the multi-tasking system to other Arabic dialects.

%\section*{Acknowledgements}

\bibliographystyle{coling}
\bibliography{coling2020}

\end{document}